\documentclass{article}
\usepackage{ijcai18}

\usepackage{natbib}

\usepackage{xcolor}
\usepackage{soul}
\usepackage[utf8]{inputenc}
\usepackage[small]{caption}

\usepackage{amsmath,amssymb,amsfonts}
\usepackage{algorithmic}
\usepackage{pgfplots}
\usepackage{graphicx}
\usepackage{textcomp}
\usepackage{clrscode3e}

\usepackage{todonotes}

\title{
Specifying and achieving goals in open uncertain robot-manipulation
domains\thanks{This paper was completed in January, 2019.}}

\author{
Leslie Pack Kaelbling \and
Alex LaGrassa\thanks{Affiliation is Carnegie-Mellon University as of
  Sept., 2019.} \and 
Tom\'as Lozano-P\'erez
 \\ 
Massachusetts Institute of Technology
 \\
{\tt 
  lpk@mit.edu,
  lagrassa@cmu.edu,
  tlp@mit.edu }
 }

\begin{document}
\maketitle

\begin{abstract}
This paper describes an integrated solution to the problem of
describing and interpreting goals for robots in open uncertain
domains.  Given a formal specification of a desired
situation, in which objects are described only by their properties,
general-purpose planning and reasoning tools are used to derive
appropriate actions for a robot.
These goals are carried out through
an online combination of hierarchical planning, state-estimation,
and execution that operates robustly in real robot
domains with substantial occlusion and sensing error.
\end{abstract}

\section{Introduction}


We would like to have intelligent robots that perform tasks in complex
open environments such as homes, warehouses, and hospitals.  As robots
become more sophisticated, tasks can be specified using high-level
goals, which the robot achieves by formulating and executing plans to
move through, sense, and manipulate the world around it.

In such domains, the goals specified by humans for the robot are
generally states of the world, rather than states of the robot,
requiring some objects in the world (dishes, boxes, medicine bottles)
to be in particular locations (a dishwasher, loading dock, or
patient's table) or states (clean, taped shut, empty).  It is critical
to be able to specify such goals even when there is substantial
uncertainty in the domain: it might be that neither the human nor the robot is
aware of the location, state, or even existence of the particular
objects needed when the goal is articulated.

\newcommand{\figwidtha}{0.35\linewidth}

\begin{figure}[t]
  \hfill
  \includegraphics[width = \figwidtha]{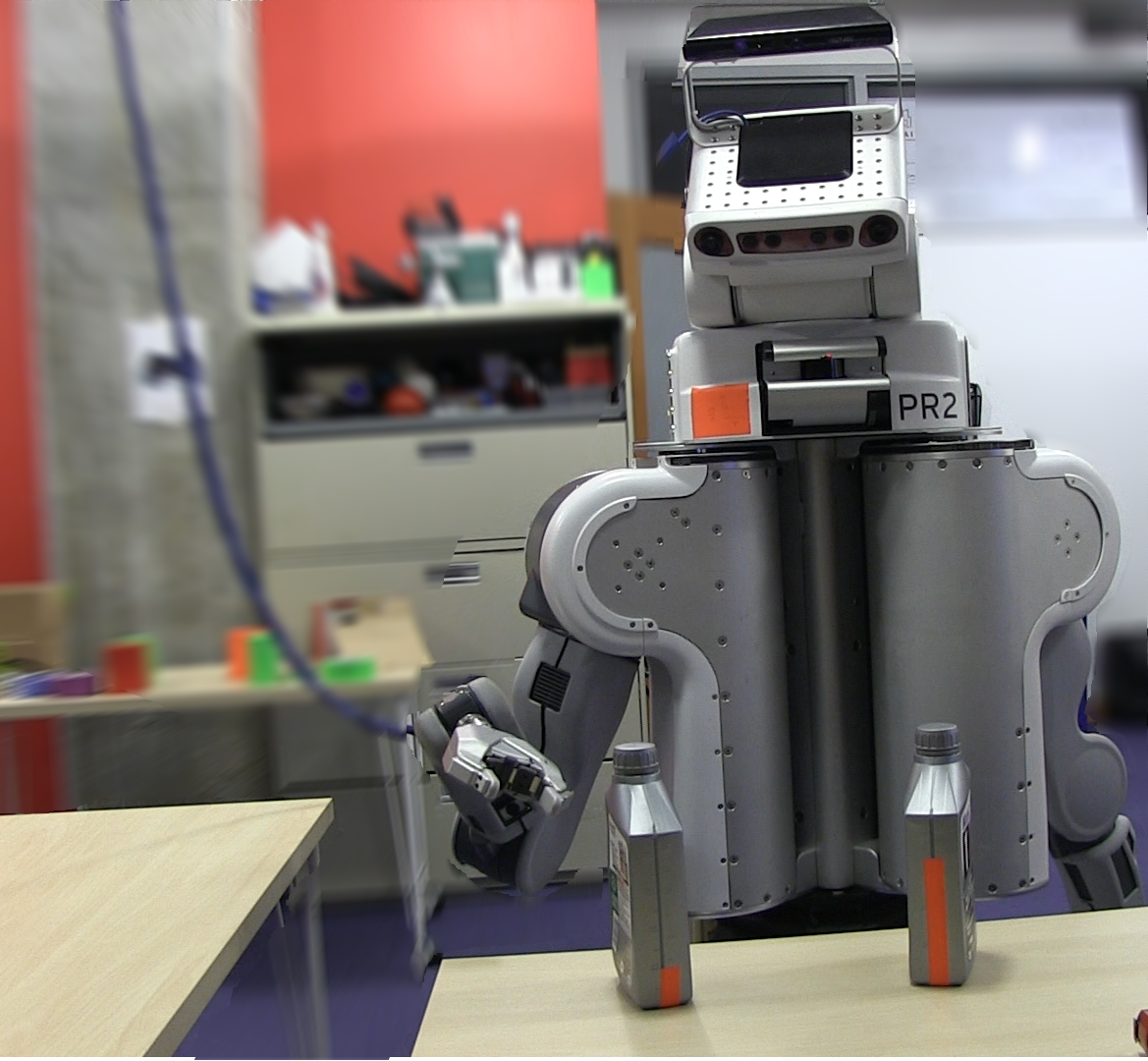}%
  \hfill
  \includegraphics[width = \figwidtha]{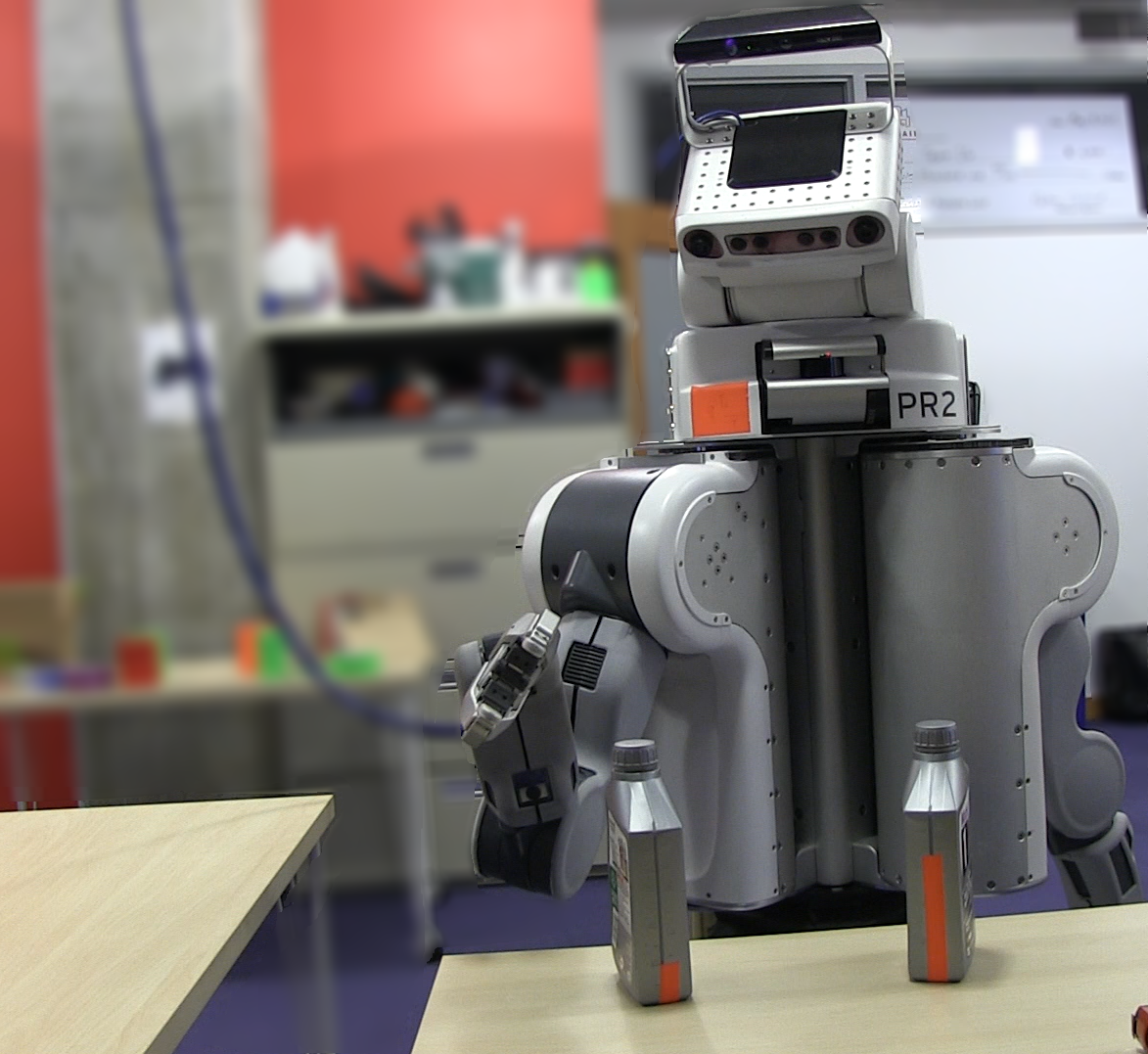}%
  \hfill\\
  \vspace{0.1cm}
  \hfill
  \includegraphics[width = \figwidtha]{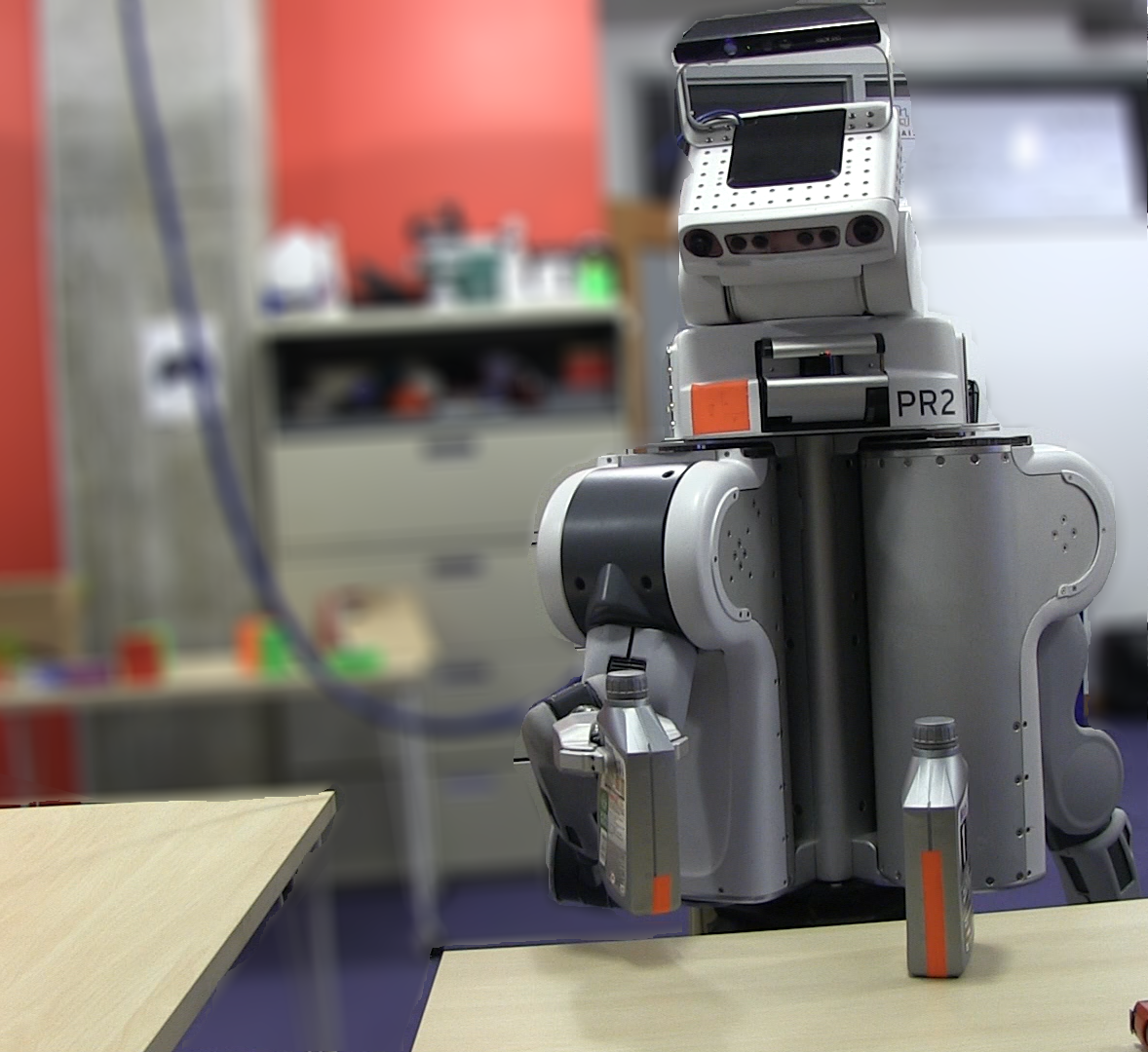}%
  \hfill
  \includegraphics[width = \figwidtha]{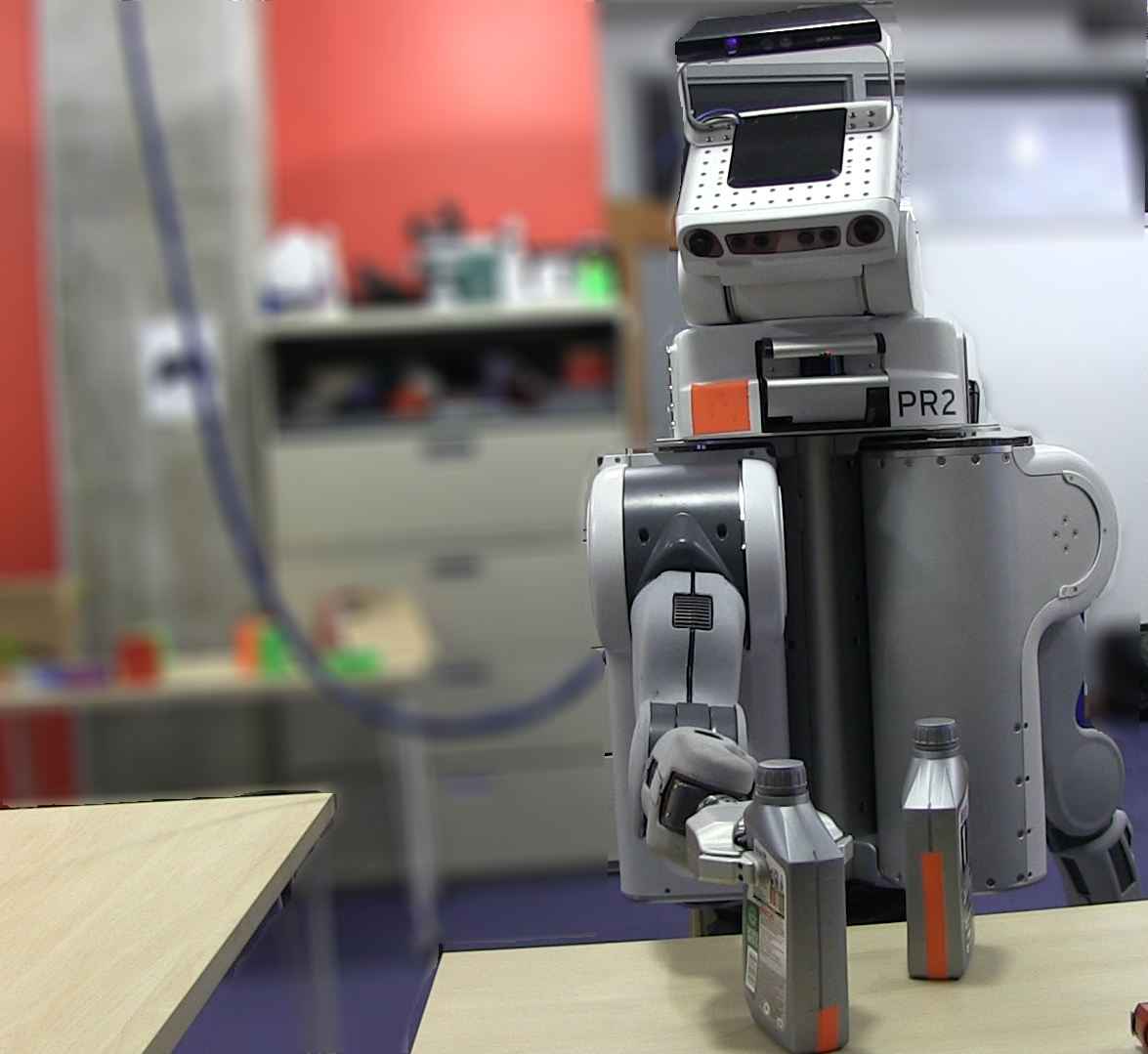}%
  \hfill\\
  \vspace{0.1cm}
  \hfill
  \includegraphics[width = \figwidtha]{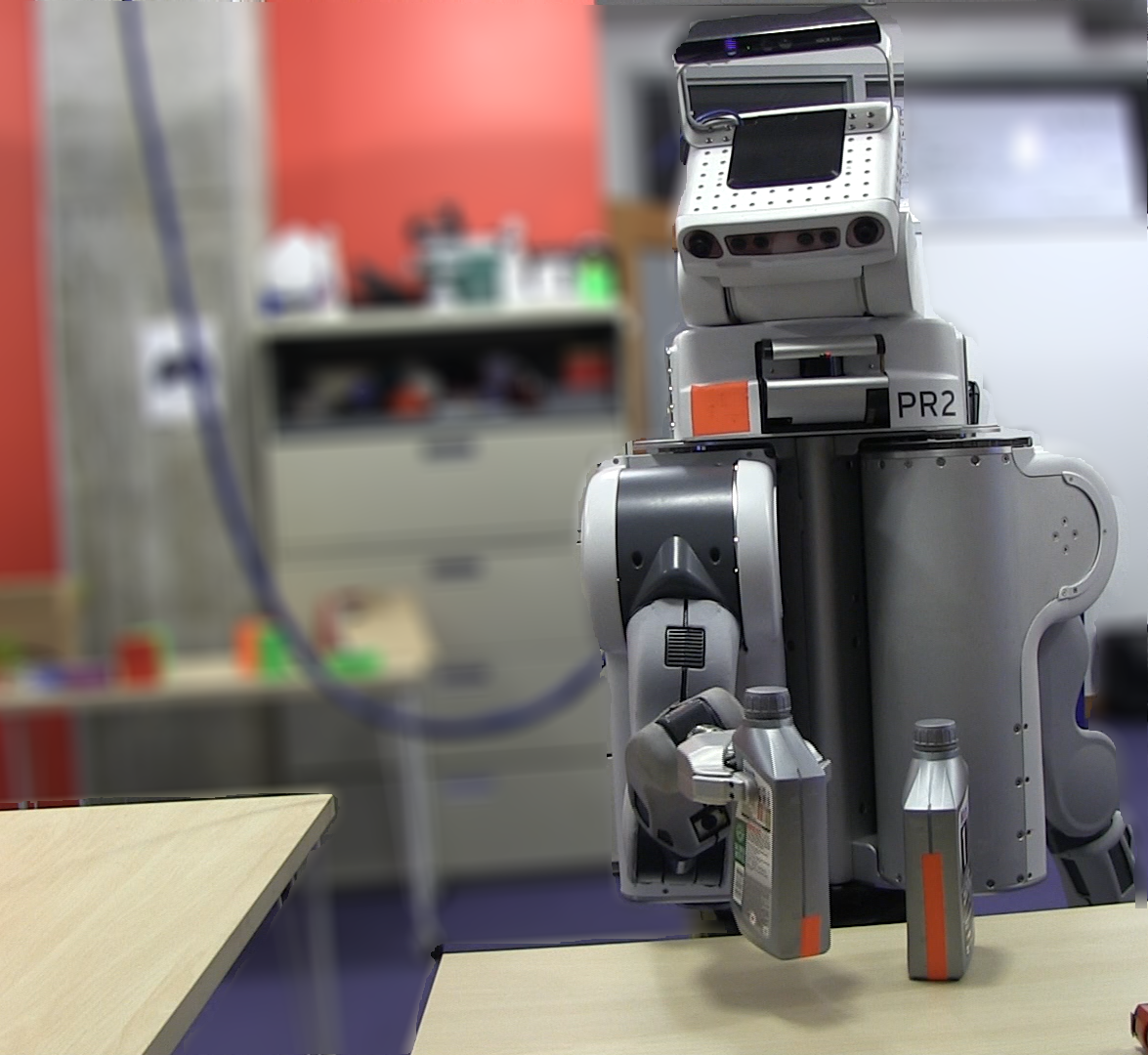}%
  \hfill
  \includegraphics[width = \figwidtha]{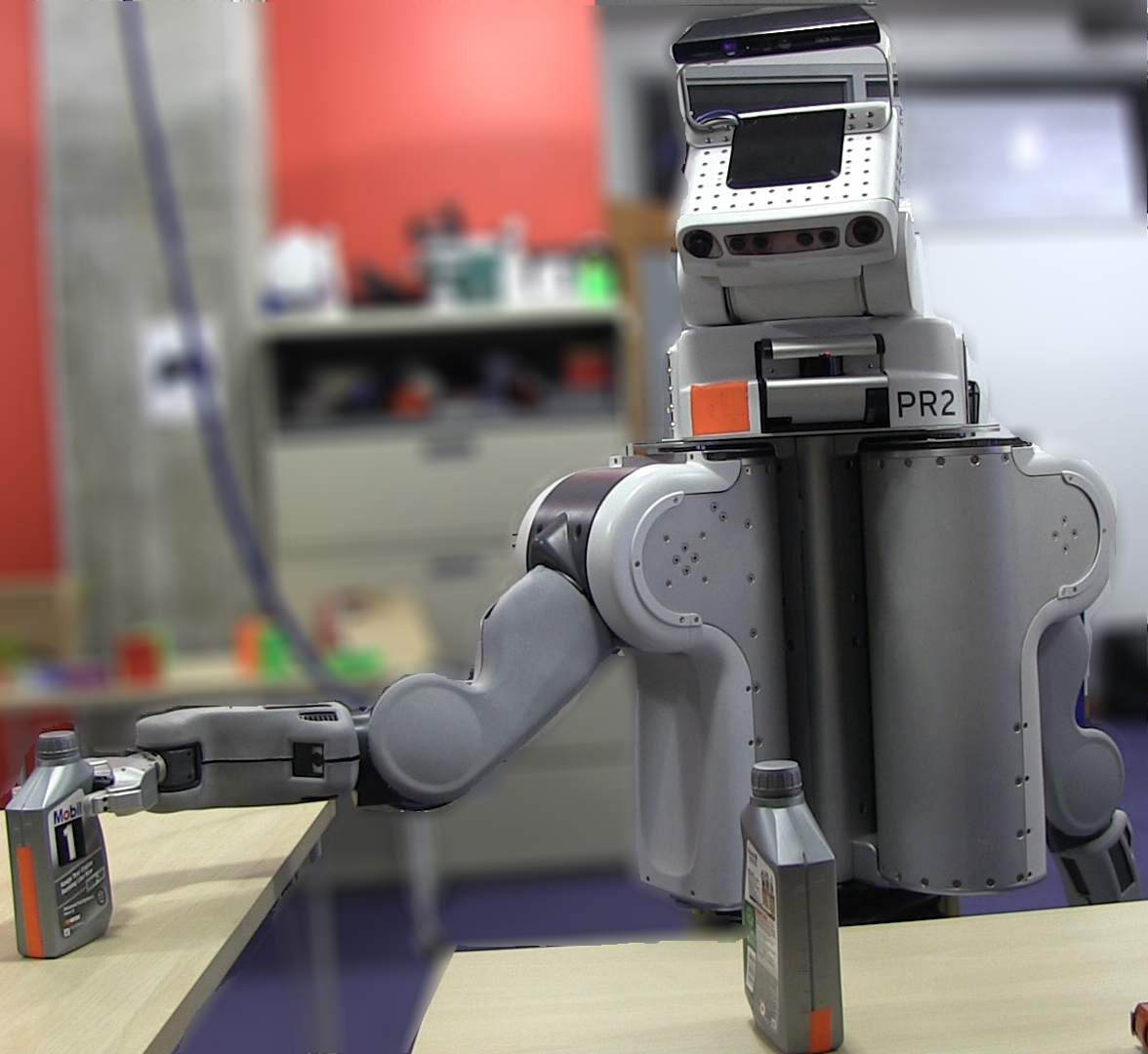}%
  \hfill\\
        \caption{Actions in response to the goal of having a heavy oil
          bottle on the table to the right.  The robot picks up one
          bottle, feels that it is light, picks up the other one, and
          places it on the correct table, requiring re-planning and 
          sensing.}
        \label{fig:executionOil}
\end{figure}

Given such a goal, the robot might have to do significant work in the
physical world just to be able to interpret it concretely: the robot
might need to search for appropriate objects or measure properties of
known objects to see if they are suitable for the task.  In this
paper, we describe an integrated solution to the problem of
describing, interpreting, and carrying out goals for robots in open
uncertain domains.  A critical feature of our approach to
understanding the meaning of goal expressions is that it is carried out
through the same planning, inference, and execution mechanisms as are
used for determining physical robot actions.  Thus, the system can use
all of its physical abilities in service of gathering information in
order to understand goal expressions in a way that will allow it to
take physical actions to achieve the ultimate objective.

In this paper, we
\begin{itemize}
\item show how to describe goals involving partially
  specified objects to a robot in an uncertain domain;
  \item provide inference
rules and planning operators that can be used to augment
an existing system for robot manipulation planning and execution under
uncertainty so that it can act to interpret and achieve these goals;
and
\item demonstrate that the robot can actively interpret goals in open
domains, by interacting with physical objects and searching the space
around it, both in simulation and on a real physical robot, in the
presence of substantial occlusion and sensing error.
\end{itemize}
Figure~\ref{fig:executionOil} provides an illustration of the
integrated system running on a PR2 robot.

Formally, this problem is a partially observed Markov decision process
({\sc pomdp}), although it would be very difficult to formalize it as
such given fundamental uncertainty even about the dimensionality of
the underlying state space.  Our solution takes inspiration from
online approximation strategies in which:
\begin{itemize}
\item goals for the system are
  specified in {\em belief space},
\item the robot makes optimistic open-loop plans in belief space,
\item after executing the first action of the plan, it
obtains an observation, updates its belief (which may
include reasoning about free space and the addition of newly
postulated objects to its representation) and replans if
necessary~\citep{Platt}.
\end{itemize}

\section{Related work}


Much of the previous work in reference resolution for robots has focused
on ambiguity in the
utterance~\citep{gorniak,schutte,khayrallah}, including purely
syntactic ambiguity as well as reference ambiguity.
\citet{tellex2011} were one of the first groups to 
symbolically ground references in natural language to concrete
representations in the world using probabilistic models. These
approaches are largely passive, in the sense that they do not
explicitly plan to gather disambiguating information.

Some planners can solve problems in open worlds, in which the 
robot does not know about all of the objects in advance.  These
methods do not explicitly
model uncertainty in the planner, which limits their ability to handle 
noisy environments, but they do have the ability to select actions to
gather information about unknown objects. 
For example, the planner used by~\citet{talama} can satisfy
quantifiable goals referring to unknown objects, like ``a human,''
but the planner does not model uncertainty about properties specified
by the goal. Furthermore, their approach does not account for actions
that modify the state of the objects the robot is sensing, which is
important for mobile manipulation domains.  Replanning occurs when the
robot discovers something new about the environment, like a new
object, but not on plan failure, which makes the
approach somewhat  less robust than ours in noisy domains.
The planner used by \citet{joshi} computes policies based on all
possible maps, which is computationally expensive. It uses a reactive
policy to avoid replanning, but must do so when a new object is
discovered in the world.  




Robots can additional use dialog with
humans to disambiguate references by detecting reference and world
ambiguity~\citep{schutte,hough2017}, and in some
work additionally determining which questions would be useful to
ask~\citep{deits,williamsdp,eppe}. For 
example, \citet{Mavridis} combined sensing and
clarification questions to resolve references with a system using
single-step lookahead and discrete properties. 

Our work focuses on the case where there is no uncertainty about the
specification, but significant uncertainty about the domain.  Planning
to disambiguate the goal specification is handled by the same
mechanism as planning to achieve goals more generally, allowing the
robot to use all of its mobile manipulation capabilities in service of
understanding and then achieving goals.




Another line of related work in the robotics community involves symbol
grounding and {\em anchoring}. The work that is closest in spirit to ours is
that of \citet{Coradeschi2001PerceptualAO}, which focuses on creating
a mapping between a symbol system and objects in a perceptual system;
they implement a system that can construct conditional plans with
observation actions to find an appropriate mapping, but it is unable
to address problems in which the plan involves objects that were not
previously known.  There are more modern extensions that have larger
scope and more general perception~\citep{Beeson,Tenorth}, and that
handle complex natural language~\citep{Lemaignan} but do not take
physical actions to aid interpretation of instructions.

\section{Denoting objects}

In order for a robot to interpret a goal in an open world, it must
\begin{itemize}
\item have its own internal beliefs about the world state,
  \item have a language
    of expressions for objects, and
    \item have a way of evaluating expressions
with respect to its current belief in order to determine which
objects it is currently aware of, if any, are likely to satisfy the
expression.
\end{itemize}

\paragraph{Belief representation and reasoning}
We assume the robot has a representation of its belief about the
world that is organized in terms of objects and probability
distributions over their properties.  Concretely, in our
running example, there are rigid objects with distributions over
properties:
\begin{description}
\item {\bf type}: multinoulli distribution over a fixed finite set of possible object types;
\item{\bf pose}: objects are assumed to be resting on a stable face, so the
pose has four degrees of freedom, $(x, y, z, \theta)$; we represent a
joint distribution over all object poses, together with the robot's
base pose, using a multivariate Gaussian in tangent space, which is
updated using a variant of the unscented Kalman filter~\citep{Hauberg}; 
\item{\bf color}: truncated
Gaussian in hue-saturation-value space;  
\item{\bf weight}: Gaussian in
log-weight space.
\end{description}
This representation is designed to support a high-fidelity
belief-update step based on object detections from a 3D
sensor for type and pose, RGB pixel values for color, and
force-torque measurements from the wrist for weight. 

In this representation, objects have no given names, unless they were
specified for the robot in advance.  
So, for example, one might
provide an initial belief describing, at least roughly, the positions
of some known objects such as tables, and give them explicit names at
initialization time.  Any objects that the robot discovers as it is
interacting with the domain, however, are added to the belief state
with internal indices that have no external meaning, but that can be
used as {\em anchors} to instantiate existentially quantified variables.

For the purposes of planning and reasoning, we characterize
sets of detailed beliefs using a language of {\em belief fluents},
which are a type of epistemic operator.  Let $\phi$ be a fluent
representing a Boolean random variable, such as whether an object is
contained in a region.  Then we will define
\[B_b(\phi, p) \equiv P_b(\phi = \mathrm{True}) \geq p\] to mean that the agent {\em
  believes} $\phi$ holds with probability at least $p$ in the current
belief state $b$, though we will generally suppress the $b$ subscript
for clarity.  It is a fluent because during execution the
underlying belief $b$ will change, and so  will the truth
value of the belief fluent.  Similarly, for a continuous random
variable, such as the color of an object, we
define 
\[B(\phi, \mu, \Sigma, \Delta, p) \equiv |\mu - M_b| \leq \Delta \wedge
  S_b \preceq \Sigma \wedge p \geq P_b\]
where the belief distribution on quantity $\phi$ can be described by a
Gaussian with parameters $M_b, S_b$ that has mixture weight $P_b$, and
where $\Sigma_1 \preceq \Sigma_2$ is a relation on covariance matrices
that holds if the equi-probability contour of $\Sigma_1$ is 
contained in the equi-probability contour of $\Sigma_2$ for any fixed
probability.  This describes a set of probability distributions that
are close (within $\Delta$) in mean and at least as certain as a
specified distribution.  

\paragraph{Denoting expressions}
Denoting expressions can be used to describe objects in terms of their
properties without explicitly naming them.  Because of uncertainty in
the underlying properties of the objects, we can never be certain
whether a denoting expression holds of a particular object; instead we
will characterize the robot's belief using belief fluents of the form:
$B(\proc{Den}(\id{expr}, \id{obj}), p)$, which means that the robot
believes, with probability at least $p$, that \id{obj} can be denoted
by \id{expr}, where \id{obj} is an internal name, or {\em anchor} for
an object.  These denoting expressions are {\it indefinite} and so it
is possible that one might be true simultaneously for many different
values of \id{obj}, or none at all.  {\it Definite} descriptions,
which imply the existence of a single satisfying object, are of great
importance, but not handled in our current implementation.

We use a variation on classical lambda expressions of the form
$\lambda X. \id{expr}$
where $X$ is a variable that may occur in \id{expr};  legal
expressions include conjunctions, disjuctions, and existential
quantification. 


The probability that random fluent $\id{Den}(\lambda X. \id{expr}, O)$
is true in a belief state $b$ is computed by recursion on
\id{expr}. 
 Let $\sigma$ be a {\em substitution} which maps variables into
constant symbols.  The application of a substitution to an expression
replaces all free occurrences of each variable in $\sigma$ with the
associated constant.  
We will write $\sigma(\id{expr})$ to stand
for the expression that results from applying substitution $\sigma$ to
\id{expr}, and write substitutions in the notation of Python dictionaries.

We assume the ability to find the probability of a {\em ground}
relational expression $\id{R}(c_1, \ldots, c_n)$ where
$c_1, \ldots, c_n$ are numeric constants or anchors to objects, in
$b$.  So, for example, we would evaluate $\id{Red}(\id{\_o34\_})$ by
finding the distribution on the color of \id{\_o34\_} in $b$ and
integrating the probability over the set of colors defined to be
red.\footnote{The semantics of color expressions in natural
  language is subtle and complex; we define
  color names as fixed volumes in HSV space.}  We will write
this quantity as $b(\id{R}(c_1, \ldots, c_n))$.  The probability that
the random fluent $\id{Den}(\lambda X. \id{expr}, O)$ is true is
$\proc{eval}(\{X : O\}(expr), b)$.  
Making strong independence assumptions, we define 
\begin{itemize}
\item $\proc{eval}(R(c_1, \ldots, c_n), b) = b(\id{R}(c_1, \ldots, c_n))$.
\item $\proc{eval}(\proc{And}(\id{expr}_1, \id{expr}_2), b)$\\ $=
  \proc{eval}(expr_1, b) \cdot \proc{eval}(expr_2, b)$.
\item 
$\proc{eval}(\proc{Or}(\id{expr}_1, \id{expr}_2), b) = \proc{eval}(expr_1, b) +
  \proc{eval}(expr_2, b) - \proc{eval}(expr_1, b) \cdot
  \proc{eval}(expr_2, b)$.
\item 
$\proc{eval}(\proc{Exists}(X, \id{expr}), b) = \proc{eval}(\bigvee_{o
  \in {\cal U}} 
  \{X : o\}(\id{expr}))$,
where ${\cal U}$ is the universe of objects.
\end{itemize}

\paragraph{Rigid designators}
In order to plan in situations when it is not initially clear which
objects will be used to satisfy a goal, we need to reason about
whether the robot concretely knows which objects it needs to
manipulate.  We can, for example, call the \id{Place} operator on any
object that is represented in the belief state using its internal
anchor as a name, but we cannot call it on $\lambda x. \id{green}(x)$
until we know of a specific object that is denoted by that expression.
In work on epistemology~\citep{KripkeNamingNecessity} and AI approaches
to planning under uncertainty~\citep{BobMoore,Morgenstern} the concept
of a {\em rigid designator} plays an important role: it is a special
name for an object (person, etc.) that always means the same thing,
independent of context, and which can be used to specify a concrete
operation.
In our formulation, internal anchors are rigid designators that can
serve as arguments for operations, but lambda expressions and
existentially quantified variables are not.  We introduce a fluent 
$\id{KRD}(A)$, where $A$ may be a variable or a
constant;  it has value \id{True} if and only if $A$ is an internal
anchoring constant.  A precondition for any operation on an object
will be that we {\it know a rigid designator} for it.  


\section{Planning and inference}

Rather than attempt to formulate and solve a {\sc pomdp} exactly, 
we follow the effective approximation
strategy, known in some circles as {\em model predictive control} and
others as {\em replanning}~\citep{Yoon}, in
which we repeatedly:
\begin{itemize}
\item Make an approximately optimal plan to achieve a goal (specified
  in belief space) given the current belief state;
\item Execute the first step of the plan; then
\item Make an observation and use it to update the belief.
  \end{itemize}
Rather than making a conditional plan, which requires branching both
on actions and observations, we plan in a determinized
model~\citep{Platt} in which it is assumed that observation actions
result in the most likely observation.  Because the goal is in belief
space (typically, to believe with high probability that some desired
world state holds) the plans will contain actions that gain
information, and if those actions result in unexpected observations,
a new plan will be made based on the new belief.

We assume a regression-based (backward) planning algorithm that uses
{\sc strips}-like rules with preconditions and results in the form of
belief fluents, and with variable values
drawn from discrete and continuous domains.  We 
augment the planner with inference rules that can be used during
backward chaining and play the role of axioms, and assume the planner
can operate hierarchically, postponing detailed planning until more
information is available.


\paragraph{Basic mechanisms}
Planning and inference rules have the form\\
$\begin{array}{ll}
\kw{precond:} & (\psi_1(\theta) = u_1), \ldots, (\psi_l(\theta) = u_l) \\
\kw{result:} & (\phi_1(\theta) = v_1), \ldots, (\phi_k(\theta) = v_k) \\
\end{array}$ \\
The $\theta$ arguments are vectors of variables; the
$\phi_i$ and $\psi_i$ are fluents that may have constants or elements
of $\theta$ as arguments; the $v_i, u_i$ are either 
constants or elements of $\theta$.  When an operator is
applied during the search, some variables in $\theta$ are bound by matching
the operator's results to fluents in the goal.  There may be
additional variables in $\theta$ that are not yet determined; these
represent the variety of ways that the operation can be carried
out to obtain the same result. 
Physical operations are accompanied by an executable
procedure, parameterized by aspects of $\theta$.  We
assign a cost to each action or inference step that is $-\log p$ where
$p$ is the probability that the action will have the desired outcome;
by finding a plan that minimizes the sum of these costs, we will have
found the open-loop plan that is most likely to achieve the
goal. 

When planning in belief space, it is typical for actions to have
belief preconditions: for example, a robot cannot attempt to pick up
an object unless its belief about the location of that object has low
variance.  However, when an object's pose is not yet well known, not
only is it impossible to pick the object up, it is impossible to plan
in detail for how to pick it up (which will depend on the object's
pose, what other objects might surround it, etc.).  For these reasons,
we assume a hierarchical planning mechanism that is able to make plans
at a high level of abstraction (by postponing preconditions in the
style of \citet{Sacerdoti}), and begin refining the
initial step and eventually taking primitive actions without planning
in detail for later parts of the plan.  This mechanism makes it
possible to delay detailed planning for physical actions, such as
picking up an object, until the high-level precondition of knowing a
rigid designator has been achieved.

\paragraph{Inference for goal interpretation}
Given these planning and reasoning mechanisms, the ability to
reason about denoting expressions and to plan and execute actions in
service of interpreting them can be implemented by defining a few new
fluent types and inference rules.  

If the robot has a subgoal of having a rigid designator for an object
that it believes is denoted by some expression \id{expr}, one way to
achieve it is by coming to believe that some particular object
\id{Obj} has the relevant properties \id{Props}.  This reasoning is 
described in the inference rule below:

\noindent \proc{ExamineObj}$(\id{expr}, \id{Props}, \id{Obj}, P_r)$:\\
$\begin{array}{ll}
\kw{precond:} & B(\func{Holds}(\id{Props},\id{Obj}), P_r),\\
              & B(\func{Den}(\id{expr}, \id{Obj}), P_p)\\
              & \func{PropsFor}(\id{expr}) = \id{Props}\\
\kw{result:} & B(\func{Den}(\id{expr}, \id{Obj}), P_r)\\
             & \func{KRD}(\id{Obj})\\
 \end{array}$
 
The function \func{PropsFor} determines which object properties would
be useful to know in order to determine the denotation of the
expression; the cost of this inference rule (log probability of its
success) depends on $P_p$, the prior probability that object \id{Obj}
has properties \id{Props}.

An alternative strategy for achieving the same subgoal, which applies
even when there is no object with a reasonable prior probability of
satisfying the expression, is to search for such an object in
regions of space that have not previously been explored:

\noindent \proc{FindObj}$(\id{expr}, \id{Region}, \id{Obj}, P_r)$:\\
$\begin{array}{ll}
\kw{precond:} & \func{BContents}(\id{Region}, P_r),\\
              & B(\func{ExistsInRegion}(\id{expr}, \id{Region}), P_p)\\
\kw{result:} & B(\func{Den}(\id{expr}, \id{Obj}), P_r)\\
             & \func{KRD}(\id{Obj})\\
 \end{array}$
 
This rule specifies that, for some region of space, if we come to know its
contents, it may yield a belief about an object that satisfies the
denotation; again the cost of the inference step is related to the log
probability that there is such an object in the region;  this
cost-based reasoning encourages the planner to select regions for
search in which an appropriate object is most likely to
occur.

Finally, we need inference rules
that connect symbolic properties, such as \id{Green} with
underlying object properties such as \id{Color}, which will allow
further inference steps to determine that in order to gather
information about the color of an object, it is necessary to look at
it. 

\section{Illustrative example}

We illustrate the close coupling of physical actions with goal
interpretation in 
an extended example, presented in
simplified form for clarity.   Assume we have a mobile manipulation
robot that can move its base and arms, pick and place objects, and look at them. 
Consider the arrangement of objects shown in Figure~\ref{scenario}(left).
The robot has already observed the objects on the table in front of
it, but is unaware of other objects in its environment. Its belief
state is (``d'' stands for distribution in the column headings, and
``low'' for low variance, meaning high uncertainty. ``prior'' means 
the robot's distribution of the weight is a prior distribution for the
weight of an arbitrary object ):

\begin{figure}
    \centering
    \includegraphics[width=0.65\linewidth]{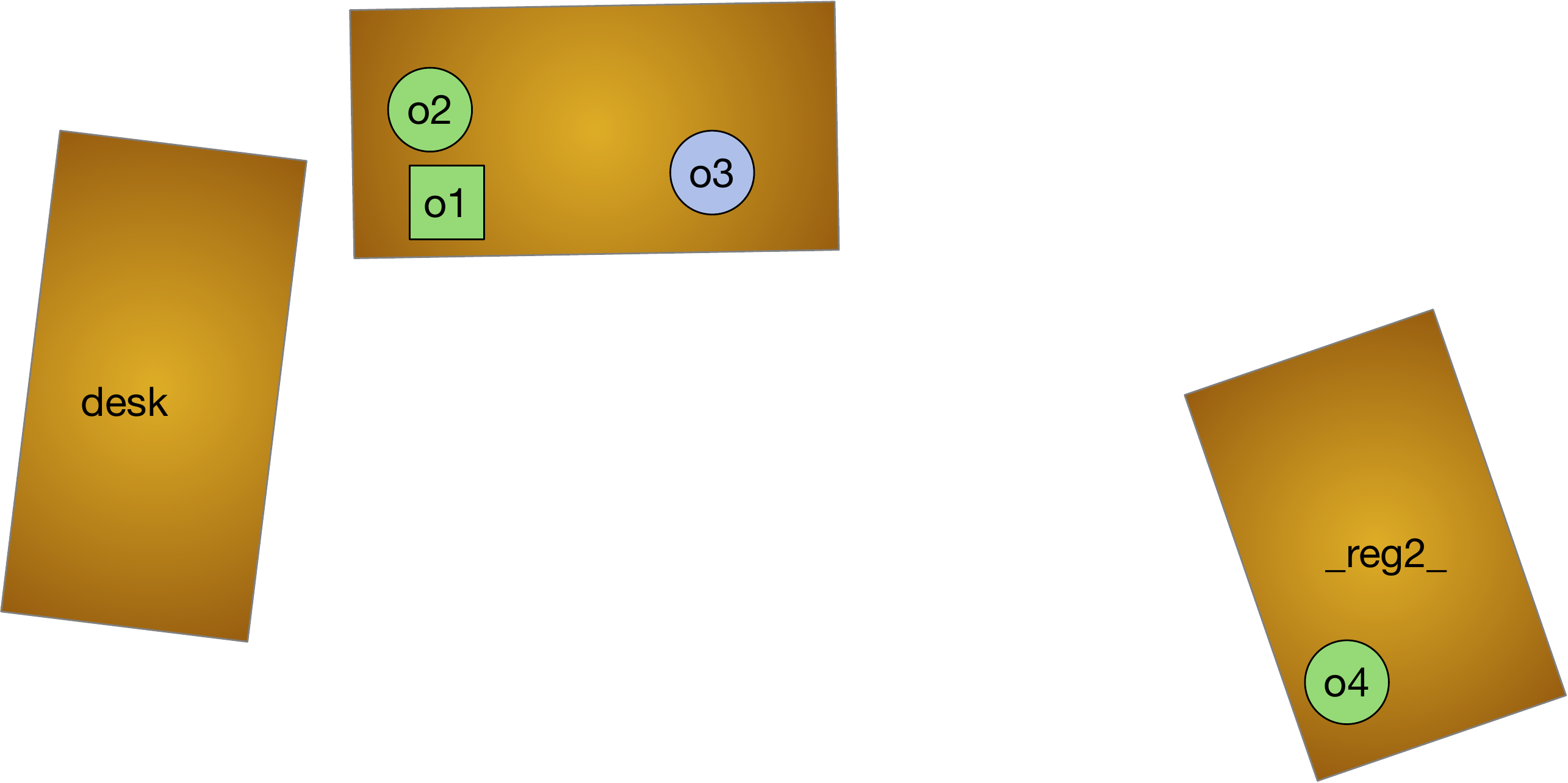}
    \includegraphics[width=0.3\linewidth]{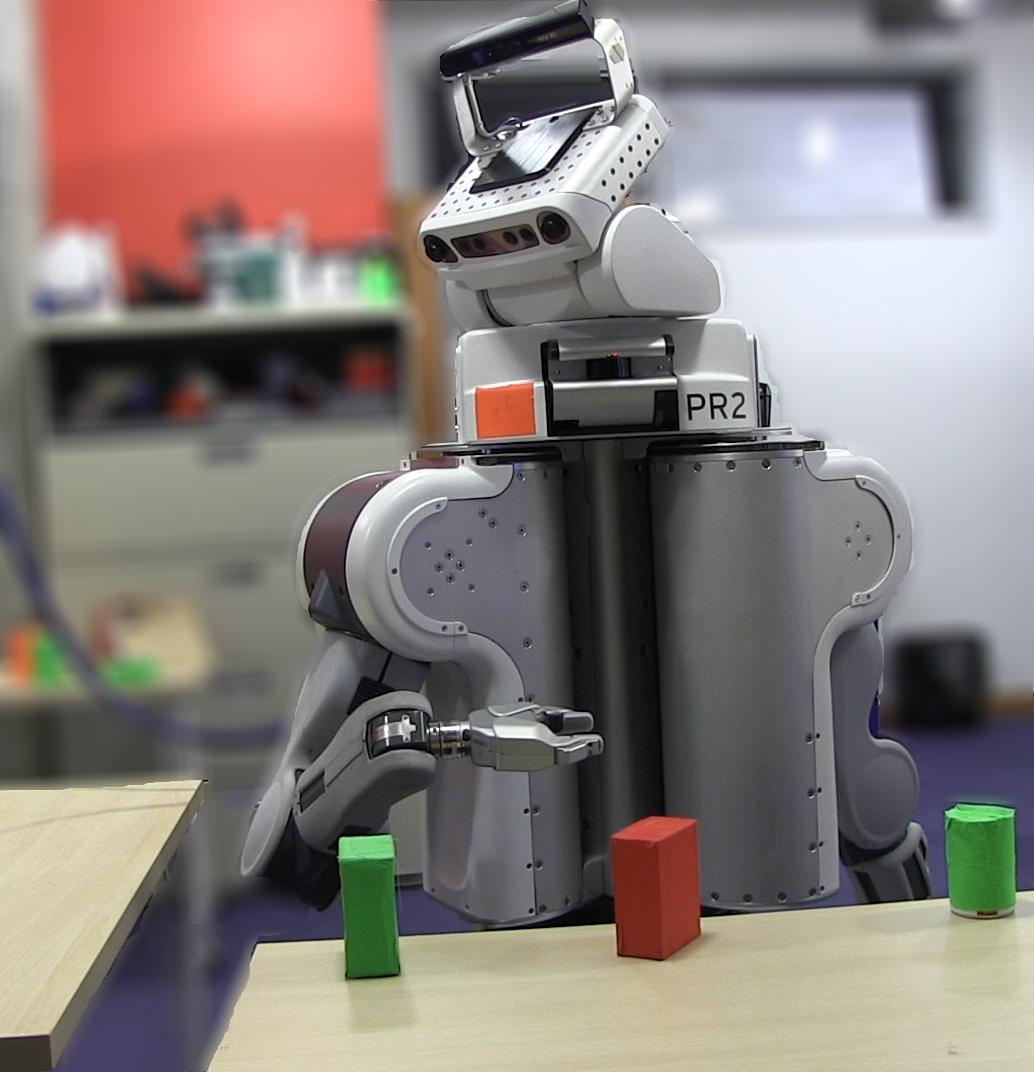}
    \caption{Scenarios for illustrative example and robot experiment}
    \label{scenario}
\end{figure}

{\footnotesize
\begin{tabular}{lllll}
\hline
id & type d & pose d & color d & weight d\\
\hline
{\tt \_o1\_} & box, .92 & (1, 1), low & green, low & prior \\
{\tt \_o2\_} & can, .80 & (2, 2), low & green, low & prior \\
{\tt \_o3\_} & can, .87 & (3, 3), low & blue, low & prior  \\
\id{desk} & table, 1.0 & (1, 1), low & brown, low & prior\\
\hline
\end{tabular}
}

\begin{figure*}[t]
    \centering
    \includegraphics[width=5.5in]{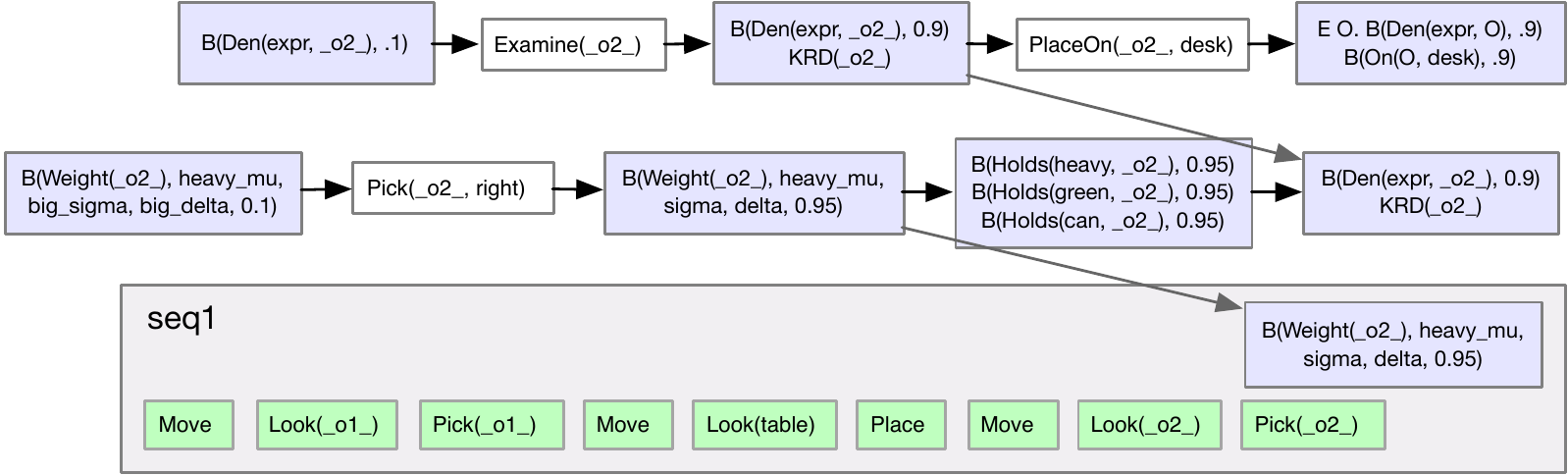}\\
\vskip0.2in
    \includegraphics[width=5.5in]{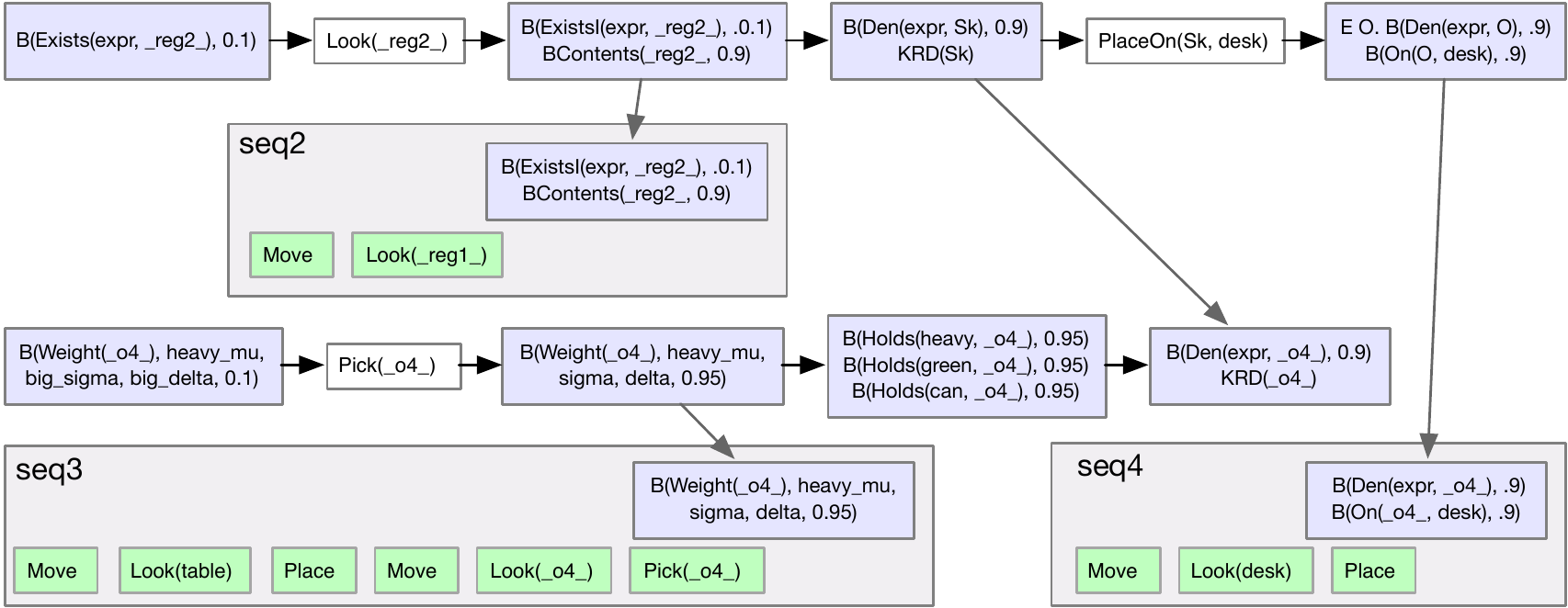}
    \caption{
Blue boxes
contain belief formulas describing goals and subgoals.  Each
horizontal row of boxes represents a plan at some hierarchical level
of the process.  Clear boxes represent the execution of an action;
arrows between blue boxes represent inference steps.  
}
    \label{examp1}
\end{figure*}

\vskip0.05in
The robot is given the goal
\[\exists o. B(\func{Den}(\id{expr}, o), .9) \wedge B(\func{On}(o,
  \id{desk}), .9) \]
where $\id{expr} \equiv \lambda x. {\id Can}(x) \wedge {\id  Green}(x) \wedge 
{\id Heavy}(x)$. 
Assume that the denotation of \id{desk} is known to the robot and the
definition of \id{Heavy} is an interval in weight space, which for
this example is objects with a mass of $400 \text{g}$ or greater.
Figure~\ref{examp1} illustrates the planning and
reasoning process.

The highest-level plan has two abstract steps:  examining an object
with the internal anchor {\tt \_o2\_}, and placing that same
object on the desk.  This plan was the most likely to succeed, which means that
the object {\tt \_o2\_} has a non-trivial probability
of being a heavy green can.  The hierarchical planning mechanism 
chooses the rightmost subgoal in that plan, and plans for it
again, but using less abstract versions of the operators with more
preconditions.   The subgoal is 
\[B(\func{Den}(\id{expr}, {\tt \_o2\_}), .9) \wedge \func{KRD}({\tt
  \_o2\_})\;\;.\]
The planner determines, through several inference steps that it should
pick up {\tt \_o2\_};  this is because it already believes with high
probability that it is green and a can, and so the weight is the
crucial property to observe; then under the assumption of most likely
observations, when it picks up the object, it will observe that it is
heavy and satisfy the goal.  Considerably more hierarchical planning,
execution, and observation results in a sequence of primitive actions,
summarized here by the first sequence of green boxes, in which it moves the
box, {\tt \_o1\_}, out of the way so that it can finally move, look at
{\tt \_o2\_} to localize its pose, and then pick it up.

The robot gets an observation that {\tt \_o2\_} weighs $100 \text{g}$ and
updates its belief state accordingly:

{\footnotesize
\begin{tabular}{lllll}
\hline
anchor & type d & pose d & color d & weight d\\
\hline
{\tt \_o1\_} & box, .92 & (10, 10), med & green, low & prior \\
{\tt \_o2\_} & can, .80 & in hand & green, low & 100, low \\
{\tt \_o3\_} & can, .87 & (3, 3), low & blue, low & prior  \\
\id{desk} & table, 1.0 & (1, 1), low & brown, low & prior\\
\hline
\end{tabular}
}
\vskip0.05in
At this point, the pre-images of both plans on the stack are no longer
true:  the robot does not believe that {\tt \_o2\_} has a significant
probability of being heavy and therefore does not believe it could
plausibly be denoted by \id{expr}.

The planner is re-invoked, resulting in the new abstract plan in the
fourth row.  This time, there are no objects that the robot knows
about that could plausibly be the denotation of \id{expr}, so the plan
is to look in some region of space (an index into a separate spatial
data structure with internal anchor {\tt \_reg2\_}) to find an object
currently named by a Skolem (placeholder) constant \id{Sk}, and then to place that
object on the desk.  The rightmost subgoal is\\
$B(\func{ExistsIn}(\id{expr}, {\tt \_reg2\_}), 0.1) \wedge
  \func{BContents}({\tt \_reg2\_}, 0.9)$\\
which is to believe that an appropriate object could plausibly be in
this region and to know its contents well.  This goal is achieved
through planning and execution of primitive actions (detailed
reasoning is elided) until the robot makes an observation of a green
object;  the pose of this new object is sufficiently different from
objects it already knows about it that 
the state estimator adds a new object, resulting in the following
belief state:

{\footnotesize
\begin{tabular}{lllll}
\hline
anchor & type d & pose d & color d & weight d\\
\hline
{\tt \_o1\_} & box, .92 & (10, 10), med & green, low & prior \\
{\tt \_o2\_} & can, .80 & in hand & green, low & 100, low \\
{\tt \_o3\_} & can, .87 & (3, 3), low & blue, low & prior  \\
\id{desk} & table, 1.0 & (1, 1), low & brown, low & prior\\
{\tt \_o4\_} & can, .91 & (6, 6), low & green, low & prior  \\
\hline
\end{tabular}
}
\vskip0.05in
A line of reasoning similar to the one we saw before
makes the object with anchor {\tt \_o4\_} most likely to be denoted by
\id{expr} and a plan is made to pick it up to observe its weight.
After the sequence of actions labeled \id{seq4} is executed (it has to
execute a place action because it is still holding {\tt \_o2\_}), the robot 
updates its belief about the weight of {\tt \_o4\_} to arrive at the
following belief state

{\footnotesize
\begin{tabular}{lllll}
\hline
id & type d & pose d & color d & weight d\\
\hline
{\tt \_o1\_} & box, .92 & (10, 10), med & green, low & prior \\
{\tt \_o2\_} & can, .80 & in hand & green, low & 100, low \\
{\tt \_o3\_} & can, .87 & (3, 3), low & blue, low & prior  \\
\id{desk} & table, 1.0 & (1, 1), low & brown, low & prior\\
{\tt \_o4\_} & can, .91 & (6, 6), low & green, low & 500, low  \\
\hline
\end{tabular}
}
\vskip0.1in
Finally, it places {\tt \_o4\_} on the desk, satisfying the 
goal. 

\section{Robot implementation}
 We have integrated these mechanisms for reasoning about denotations
with the pick-and-place capabilities of a PR2 robot for mobile
manipulation, using the {\sc bhpn}~\citep{bhpn} planning and execution
mechanism. 
The robot has a base, two arms and head, with total of 20 DOF.
A Kinect sensor
generates colored point clouds that are used for detecting
objects;  detections are categorized by type 
and are accompanied by a color observation, computed as
the mean of the colors of the points associated to the object by the
detector.  The right wrist has a (very noisy) 6-axis
force-torque sensor, which generates indirect observations of 
the weight of the object the robot is holding.

\paragraph{Simulation}
Figure~\ref{simEx1Stills} illustrates the first scenario, with
three objects on three tables, arranged so that at most one of the
objects is in the field of view at a time.  The goal is
\[\exists o. B(\id{Den}(\lambda x. \id{Green}(x), o), 0.9)\wedge
B(\id{In}(o, \id{table1}), 0.9)\;\;.\]
Table 1 is the table directly in front of the robot.  The robot's
initial belief includes the existence of the objects, but not their color.
The robot plans to determine that the object \id{sodaE}
  satisfies the denoting expression and to place it into the region.
It observes the object, receives a color
  observation, and performs a belief update.  The new belief (that
 \id{sodaE} is probably red) means that the belief 
  state is not in the 
  pre-image of any of the plans on the stack.
The robot replans
  and finds that object \id{sodaB} is the most likely to satisfy the
  denoting expression, and so it moves and looks again, discovers that
  \id{sodaB} is blue, and pops the plan stack once more.
  It tries once more, with object \id{sodaC}, 
  discovers that it is green, then formulates and executes
  plans for picking up the object, moving, and placing it in the
  target region.

In Figure~\ref{simEx2Stills}, we begin with the same initial belief state,
but the goal is now
\begin{eqnarray*}
\exists o. & B(\id{Den}(\lambda
x.\proc{And}(\id{Heavy}(x),\id{Soda}(x)), O), 0.9))  \\ 
&  \wedge  B(\id{In}(o, \id{table1}), 0.9)
  \end{eqnarray*}
\newcommand{\figWidth}{0.23\linewidth}
\newcommand{\figHSpace}{0.1cm}
\begin{figure}
         \centering
\frame{\includegraphics[width = \figWidth]{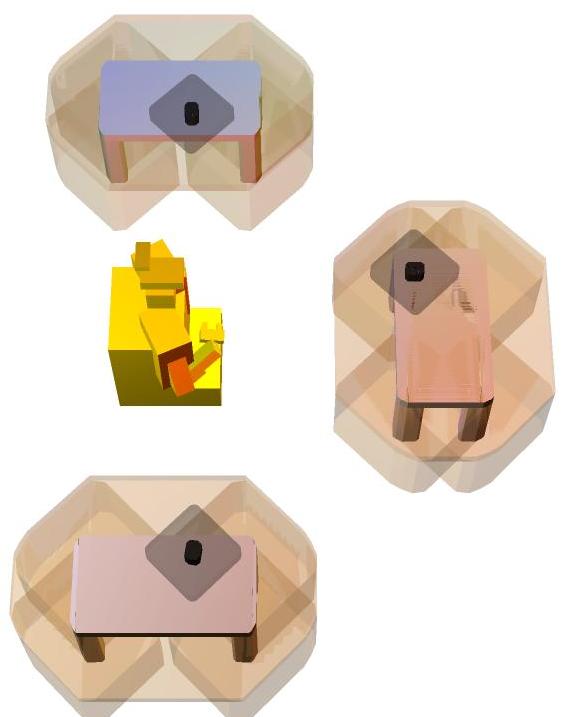}}
\hfill
\frame{\includegraphics[width = \figWidth]{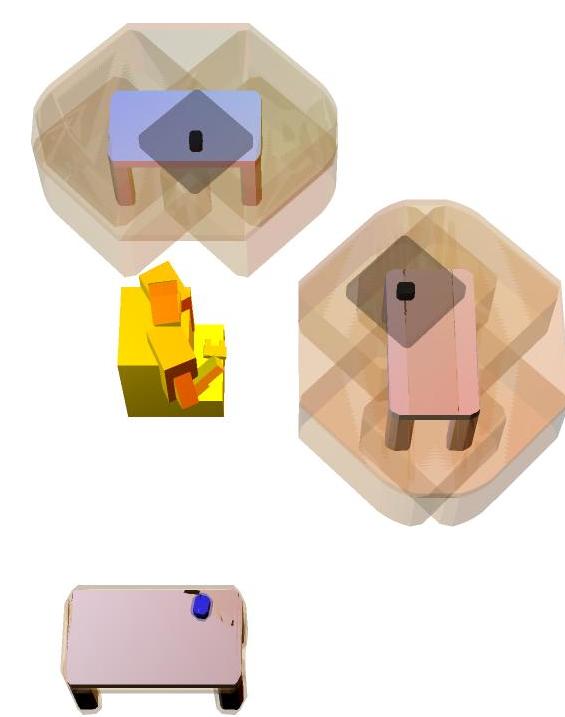}}
\hfill
\frame{\includegraphics[width = \figWidth]{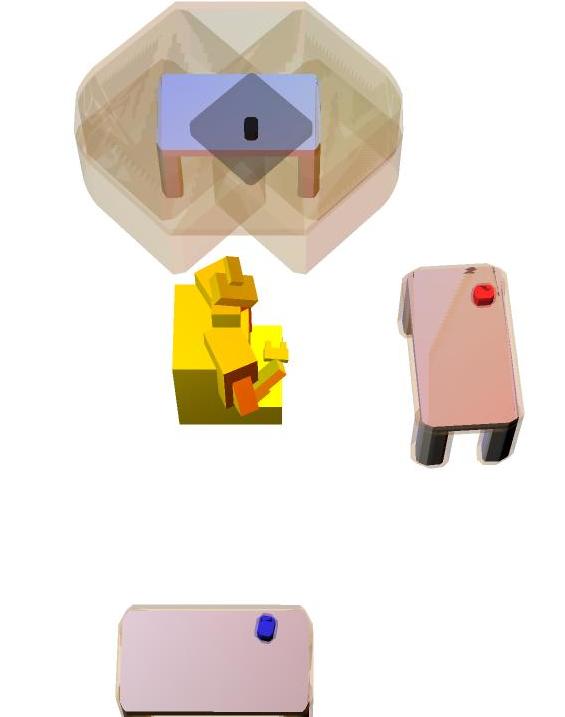}}
\hfill
\frame{\includegraphics[width = \figWidth]{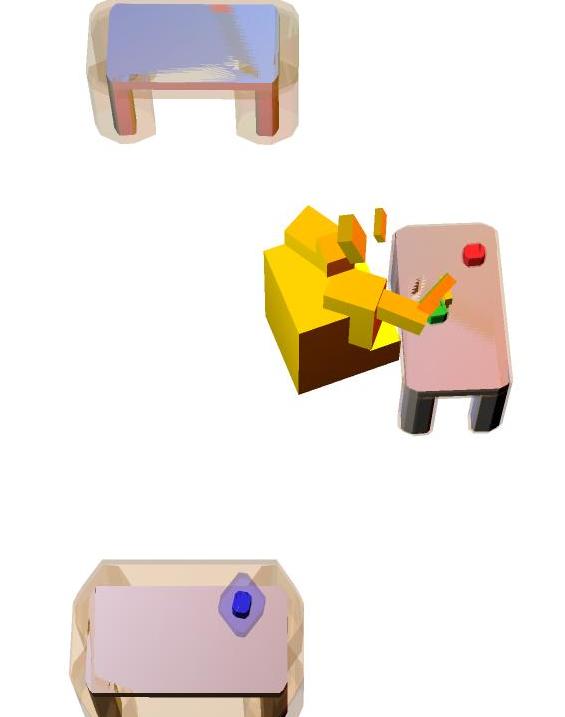}}
\caption{Goal: a green object on right-hand table.}
\label{simEx1Stills}

\frame{\includegraphics[width = \figWidth]{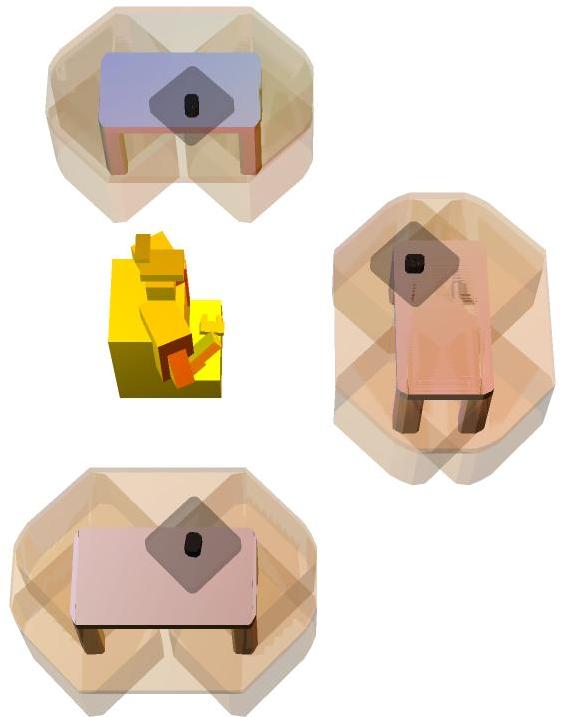}}
\hfill%
\frame{\includegraphics[width = \figWidth]{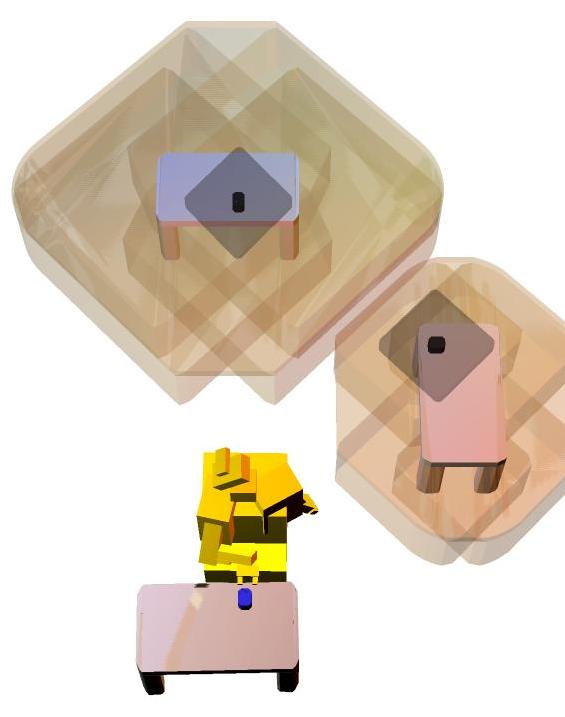}}
\hfill
\frame{\includegraphics[width = \figWidth]{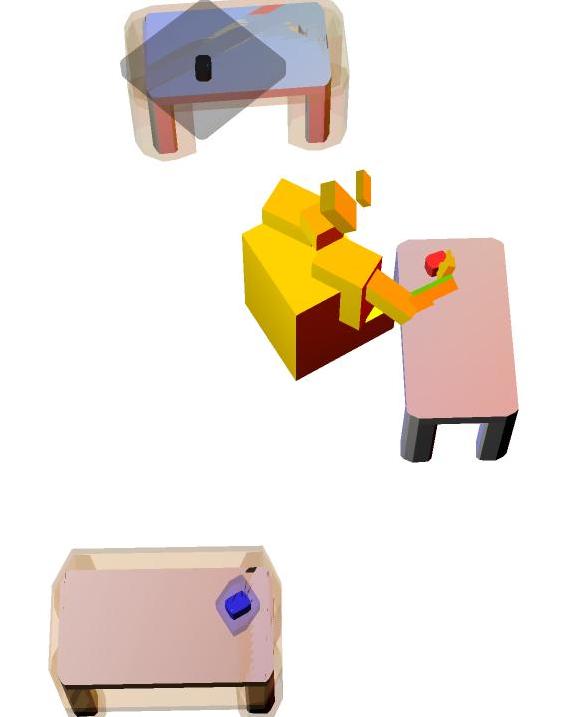}}
\hfill
\frame{\includegraphics[width = \figWidth]{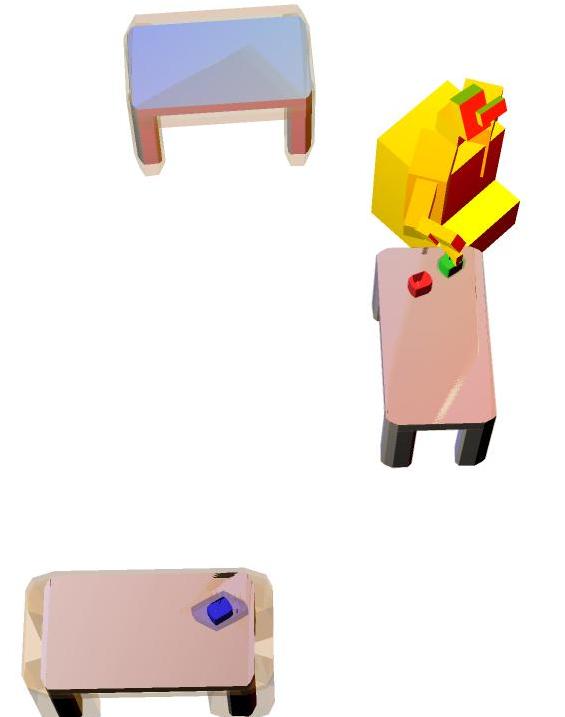}}
        \caption{Goal: a heavy object on right-hand table.}
         \label{simEx2Stills}

\frame{\includegraphics[width = \figWidth]{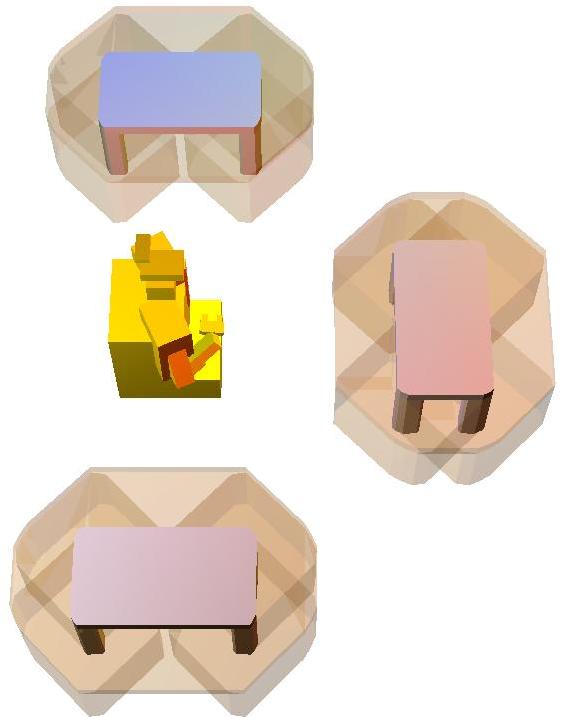}}
\hfill
\frame{\includegraphics[width = \figWidth]{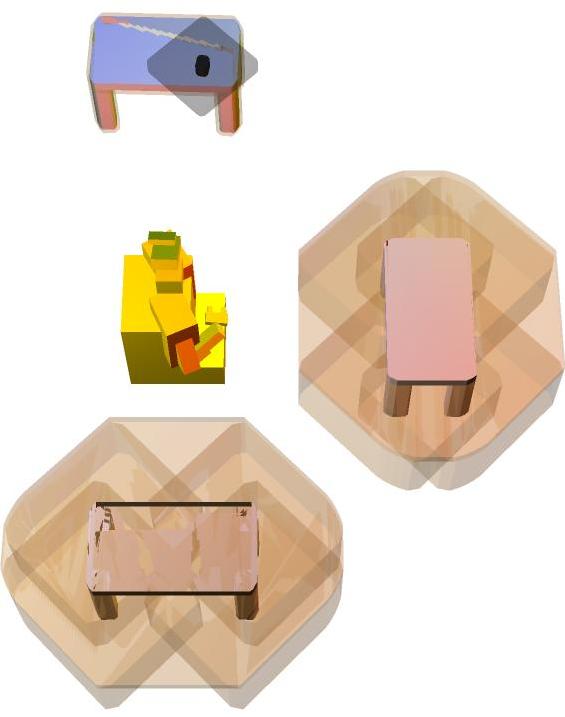}}
\hfill%
\frame{\includegraphics[width =
  \figWidth]{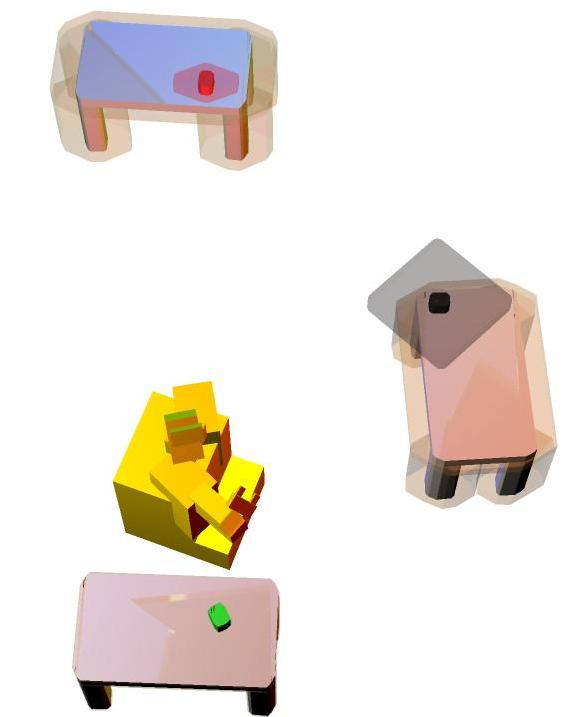}}
\hfill
\frame{\includegraphics[width = \figWidth]{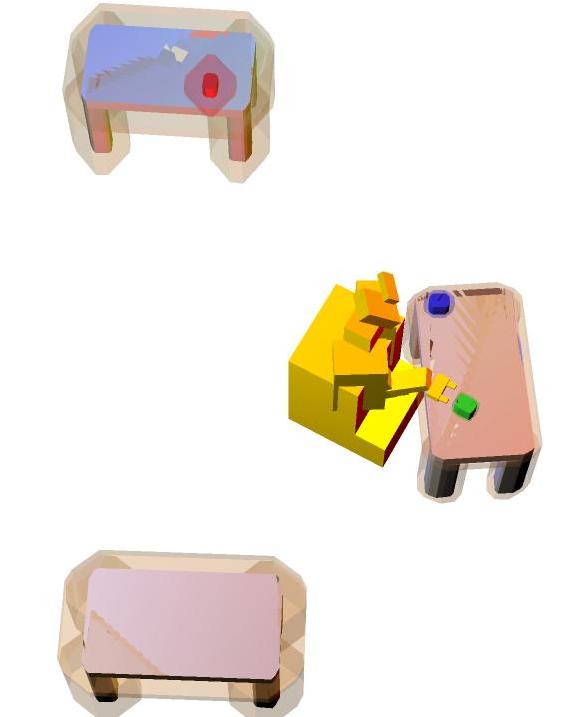}}
        \caption{Goal: a green object on right-hand table; no objects
          known in advance}
         \label{simEx3Stills}
\end{figure}

This execution has a similar structure, in which the robot
examines two objects before finding a third one satisfactory and
putting it into the target region.  However, now, 
because the objective is to find a heavy object, the robot must move over to each object in turn, and pick it up to weigh it.  Once it has
discovered a heavy object, the rest of the execution is actually
simpler, because the robot finds that it is already holding the object
it needs to place.

Finally, in Figure~\ref{simEx3Stills}, we illustrate a situation in which 
the robot knows of the existence of tables in advance, and believes
that objects are likely to be found on top of tables.  Its goal is the
same as in the first example:
$\exists o. B(\id{Den}(\lambda x. \id{Green}(x), o), 0.9) \wedge 
B(\id{In}(o, \id{table1}), 0.9)$.
The structure of the planning and execution that leads to the goal is
almost exactly the same as for the first example, except that it uses
the \proc{LookAtRegion} operator to search in regions for objects;
when it has thoroughly examined one region and not found any objects
that satisfy the denoting expression, it pops the planning stack up as
before.

\paragraph{Real robot}
The same implementation runs on the physical robot.
The first scenario corresponds to Figure~\ref{fig:executionOil}, in
which there are two oil bottles on a table, one
mostly filled with oil, and one mostly empty. The initial poses of the
bottles and tables were given with standard deviation $0.1 \text{m}$.
The robot was asked to achieve the goal
\[\exists o. B(\id{Den}(\lambda x. \id{Heavy}(x), o), 0.9) \wedge
B(\id{In}(o, \id{table1}), 0.9)\;\;, \]
which requires locating the heavy object and moving it to the table to
its right.  In this particular execution, 
the robot first looks at the oil bottle 
to its right to reduce pose uncertainty,  then it picks it up to
determining its weight.
Based on the observation, the robot updates its belief about the
weight of that object, and now believes, with high probability that
that bottle is not heavy.  It replans, and decides to pick up the
other bottle.  It is nearly full, and therefore heavy enough to
satisfy the specification. Thus, the robot places it on the table to the
right.

We repeated the experiment multiple times, varying the poses of the
oil bottles, including switching their left-to-right order.  Due to
noise in detections and nondeterminism in the planner, we observed a
variety of successful execution sequences with multiple look
operations and different order of picking the bottles, which
highlights the flexibility enabled by integrating the reference
resolution with the physical planning.
We also ran an experiment with the goal
\[
\exists o. B(\id{Den}(\lambda x. \id{Green}(x), o), 0.9)
             \wedge 
B(\id{In}(o, \id{table1}), 0.9)\;\; \]
using colored point clouds. 
The robot was given a green can, a green soda box, and a red box on
a table in front of it, and had to move a green object to the 
right table (Figure~\ref{scenario}). 
The robot reliably picks up the green soda box and moves it to 
the other table, even in the presence of rearrangement of the objects
and of pose and type errors in the
perception system, requiring different sequences of looking and motor
operations.  Videos of simulation and real robot experiments are
available at\\ {\tt\scriptsize
https://sites.google.com/view/specifying-and-achieving-goals/home}.


\paragraph{Conclusions}  It is critical to be able to communicate
goals to robots in terms of object properties even when particular
relevant objects are not known to the robot or the human.
We have demonstrated that this capability can be achieved in a 
robust and flexible way through tight integration with state
estimation and belief-state planning mechanisms that control the
robot's physical and information-gathering actions.

\bibliographystyle{named}
\bibliography{arr_icra,ijcai18}

\end{document}